\documentclass[11pt]{article}

\usepackage[preprint]{acl}

\usepackage{times}
\usepackage{latexsym}

\usepackage[T1]{fontenc}

\usepackage[utf8]{inputenc}

\usepackage{microtype}

\usepackage{inconsolata}

\usepackage{graphicx}

\usepackage[utf8]{inputenc}
\usepackage[T1]{fontenc}
\usepackage{algorithm}
\usepackage{algorithmic}
\usepackage{graphicx}
\usepackage{booktabs}
\usepackage{multirow}
\usepackage{amsmath}
\usepackage{amsfonts}
\usepackage{enumitem}
\usepackage{url} 
\usepackage{xspace}
\usepackage[table]{xcolor}
\usepackage{hyperref}
\usepackage{epigraph}
\usepackage{amsthm}
\usepackage{amssymb}
\usepackage{arydshln}
\usepackage{bbm}
\usepackage{wrapfig}
\usepackage[most]{tcolorbox}

\definecolor{langlightblue}{rgb}{0.3, 0.65, 1}
\definecolor{langblue}{rgb}{0, 0.4, 0.8}
\definecolor{langmildblue}{rgb}{0.0, 0.45, 0.73}
\definecolor{langdarkblue}{rgb}{0.0, 0.0, 0.61}
\definecolor{langred}{rgb}{0.81, 0.09, 0.13}
\definecolor{langlightgreen}{rgb}{0.80, 0.97, 0.85}
\definecolor{langgreen}{rgb}{0.18, 0.55, 0.34}
\definecolor{langdarkgreen}{rgb}{0.0, 0.45, 0.38}
\definecolor{bingpink}{rgb}{1.0, 0.41, 0.71}
\hypersetup{
    colorlinks=true,
    linkcolor=black,
    citecolor=langmildblue,
    urlcolor=langmildblue,
}
\newcommand{\gain}[1]{\textbf{\textcolor{langdarkgreen}{#1}}}
\newcommand{\loss}[1]{\textbf{\textcolor{langred}{#1}}}

\makeatletter
\AtBeginDocument{%
  \let\oldref\ref
  \renewcommand{\ref}[1]{%
    \hyperref[{#1}]{\underline{\oldref{#1}}}%
  }%
}
\makeatother

\usepackage{titletoc}

\newtcblisting{promptbox}[2][]{
    enhanced,
    width=\linewidth,
    colback=gray!5,
    colframe=gray,
    title=\textbf{#2},
    listing only,
    listing options={
        basicstyle=\normalfont\itshape\small,
        breaklines=true,
        breakatwhitespace=true,
        columns=fullflexible
    },
    #1
}

\title{PSE: Analyzing Long-Horizon LLM Agents by Decomposing Tool Use into Planning, Selection, and Execution}

\author{
Lang Cao\thanks{Equal contribution.} \quad
Hanbing Liu\footnotemark[1] \\
Independent Researchers \\
\texttt{windszzlang@gmail.com \quad hanbingliu0@gmail.com}
}

\begin{document}
\maketitle


\begin{abstract}
LLM agents solve complex tasks through long-horizon interactions with APIs and tools. Existing evaluations primarily measure end-to-end task success or assess individual tool-use capabilities separately, leaving their downstream effects on task completion less understood. Meanwhile, growing efforts to improve tool use across complex tasks highlight the need to identify which capabilities most affect overall performance. We introduce PSE, a controlled evaluation framework that decomposes the agent loop into three atomic capabilities: tool planning, tool selection, and tool execution. We substitute a weaker LLM in one component at a time while keeping the others fixed and evaluate final task success. Our results show that replacing the planning component causes the largest overall decline in task success, suggesting that tool planning is a key performance bottleneck. We further observe substantial performance gains from adding domain instructions, suggesting that effective tool planning is closely related to an LLM's understanding of domain-specific workflows. These findings suggest directions for future LLM agent development and optimization.
\end{abstract}


\section{Introduction}

LLM agents solve complex tasks through long-horizon interactions with APIs and other tools. By interleaving reasoning, actions, and observations, agents can gather information, modify external states, and coordinate multiple tools to accomplish user goals \cite{yao2023react,schick2023toolformer,qin2024toolllm}. Successful task completion requires more than generating valid API calls: agents must decide what to do next, select an appropriate tool, and construct its arguments using information obtained from previous interactions.

\begin{figure}[t]
    \centering
    \includegraphics[width=\linewidth]{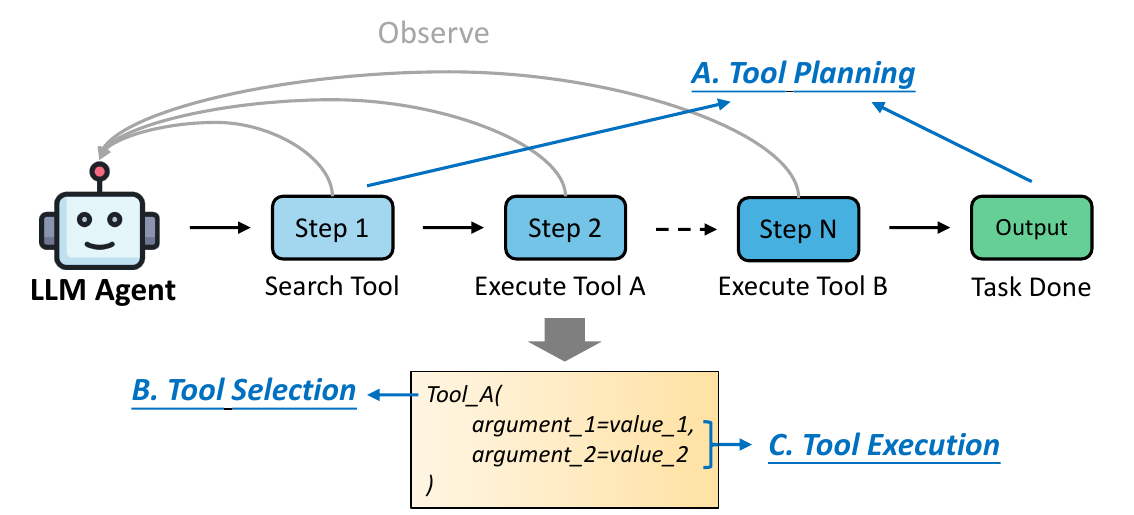}
    \caption{We decompose long-horizon LLM tool use into planning, selection, and execution. At each step, the agent observes previous outcomes and plans its next action. Tool selection determines which tool to call, while tool execution specifies the arguments for that call. The process repeats until the task is complete.}
    \label{fig:pse-teaser}
\end{figure}

Recent work improves these decisions through modular architectures and structured tool planning. HuggingGPT and RestGPT explicitly separate planning, selection, and execution \cite{shen2023hugginggpt,song2023restgpt}, while GTool and ToolTree strengthen planning through dependency modeling and trajectory search \cite{chen2026gtool,yang2026tooltree}. As long-horizon agents become increasingly prevalent, these developments raise an important evaluation question: \emph{which tool-use capabilities most constrain task success in long-horizon workflows?} Answering this question can help identify priorities for further improvement.

Existing evaluations provide only a partial answer. Interactive benchmarks measure whether agents successfully complete tasks across diverse environments and application workflows \cite{liu2024agentbench,trivedi2024appworld,barres2025tau2,shepard2026automationbench}. Other benchmarks distinguish API calling, retrieval, and planning, or evaluate task decomposition, tool selection, and parameter prediction \cite{li2023apibank,shen2024taskbench}. However, overall success rates do not isolate the effects of individual components, while component-level accuracy does not directly capture their downstream effects during interactive task execution. A planning error can lead to an inappropriate tool search, a selection error can direct execution toward the wrong API, and an argument error can change the environment state and affect subsequent decisions. These dependencies motivate us to investigate which capabilities are most critical to successful long-horizon tool use.

In this paper, we introduce \textsc{PSE}, a controlled evaluation framework that decomposes the agent loop into three atomic capabilities: \emph{tool planning}, \emph{tool selection}, and \emph{tool execution}. Planning determines whether to search for tools, execute an available tool, or finish the task, reflecting the agent's ability to guide the workflow at a high level. Selection identifies a concrete API from the available tool list, while execution constructs the arguments for the selected API, translating the plan into a concrete tool call. As illustrated in Figure~\ref{fig:pse-teaser}, these decisions recur throughout the interaction as new observations become available. Using AutomationBench \cite{shepard2026automationbench} across six domains, we assign a reference LLM to all three components and substitute a weaker LLM in one component at a time, keeping the others fixed. Each configuration begins from the same initial task state but follows its own actions and observations rather than replaying a fixed trajectory. We retain final task success as the evaluation metric, capturing how component substitutions propagate through the agent loop, including their effects on subsequent decisions and recovery.

Our contribution is a controlled, component-level evaluation of long-horizon tool use that connects model substitutions to final task success, capturing their downstream effects throughout the agent loop. By replacing one component at a time while keeping the others fixed, we find that substituting a weaker model for planning causes the largest overall performance decline, identifying planning as a key bottleneck in the evaluated settings. We further observe substantial gains from domain instructions, highlighting the importance of understanding domain-specific workflows. Together, these findings identify priorities for future LLM agent development and optimization.

\section{Related Work}

\paragraph{Tool-Augmented LLM Agents.}
External tools extend LLM agents beyond text generation. ReAct interleaves reasoning with tool interactions \cite{yao2023react}, while Toolformer, Gorilla, and ToolLLM improve API invocation through learning and retrieval \cite{schick2023toolformer,patil2024gorilla,qin2024toolllm}. Modular approaches explicitly structure tool use: HuggingGPT separates planning, selection, and execution \cite{shen2023hugginggpt}, and RestGPT coordinates a planner, API selector, and executor \cite{song2023restgpt}. Although the ReAct paradigm remains widely adopted, recent work explores more structured tool planning. GTool models dependencies using request-specific tool graphs \cite{chen2026gtool}, while ToolTree explores tool-use trajectories through tree search and bidirectional pruning \cite{yang2026tooltree}. This trend highlights growing efforts to improve agents' tool planning for domain-specific workflows. Complementing these advances, PSE provides a controlled evaluation framework examining how planning, selection, and execution affect task success and identifying priorities for future LLM agent development.

\begin{table*}[t]
    \centering
    \resizebox{0.9\textwidth}{!}{%
        \begin{tabular}{lccccccc}
            \toprule
            \textbf{Configuration} & \textbf{Sales} & \textbf{Marketing}
            & \textbf{Operations} & \textbf{Support} & \textbf{Finance}
            & \textbf{HR} & \textbf{Overall} \\
            \midrule
            \rowcolor{gray!15}
            \raisebox{-0.23em}{%
                \includegraphics[width=1.3em]{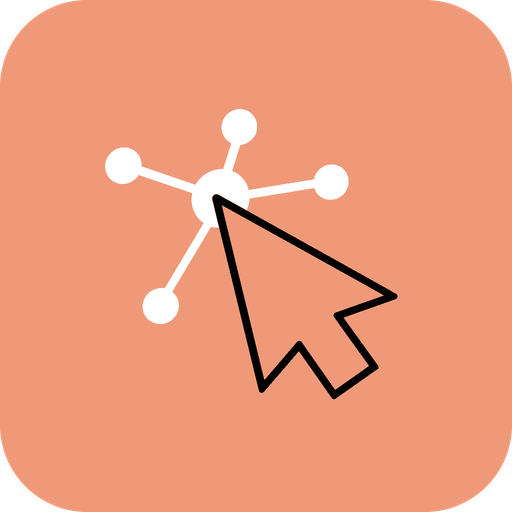}%
            }\ \textbf{Claude-Opus-5.5}
            & 0.41 & 0.39 & 0.42 & 0.15 & 0.31 & 0.28 & 0.33 \\
            \midrule
            \quad Haiku $\rightarrow$ Tool Planning
            & 0.21 & 0.21 & 0.29 & 0.08 & 0.24 & 0.12
            & 0.19 \loss{(-42.4\%)} \\
            \quad Haiku $\rightarrow$ Tool Selection
            & 0.30 & 0.31 & 0.34 & 0.11 & 0.28 & 0.22
            & 0.26 \loss{(-21.2\%)} \\
            \quad Haiku $\rightarrow$ Tool Execution
            & 0.24 & 0.22 & 0.31 & 0.11 & 0.29 & 0.19
            & 0.23 \loss{(-30.3\%)} \\
            \midrule
            \quad Sonnet $\rightarrow$ Tool Planning
            & 0.35 & 0.33 & 0.38 & 0.11 & 0.29 & 0.23
            & 0.28 \loss{(-15.2\%)} \\
            \quad Sonnet $\rightarrow$ Tool Selection
            & 0.40 & 0.37 & 0.40 & 0.13 & 0.33 & 0.26
            & 0.32 \loss{(-3.0\%)} \\
            \quad Sonnet $\rightarrow$ Tool Execution
            & 0.35 & 0.37 & 0.40 & 0.11 & 0.31 & 0.27
            & 0.30 \loss{(-9.1\%)} \\
            \midrule
            \quad w/ Domain Instruction
            & 0.63 & 0.68 & 0.53 & 0.49 & 0.60 & 0.35
            & 0.55 \gain{(+66.7\%)} \\
            \midrule
            \rowcolor{gray!15}
            \raisebox{-0.23em}{%
                \includegraphics[width=1.3em]{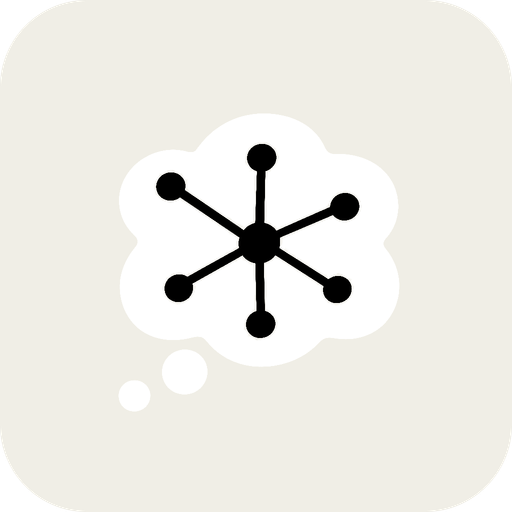}%
            }\ \textbf{Claude-Sonnet-5.5}
            & 0.29 & 0.28 & 0.39 & 0.10 & 0.24 & 0.23 & 0.26 \\
            \quad w/ Domain Instruction
            & 0.60 & 0.69 & 0.57 & 0.45 & 0.63 & 0.26
            & 0.53 \gain{(+103.8\%)} \\
            \midrule
            \rowcolor{gray!15}
            \raisebox{-0.23em}{%
                \includegraphics[width=1.3em]{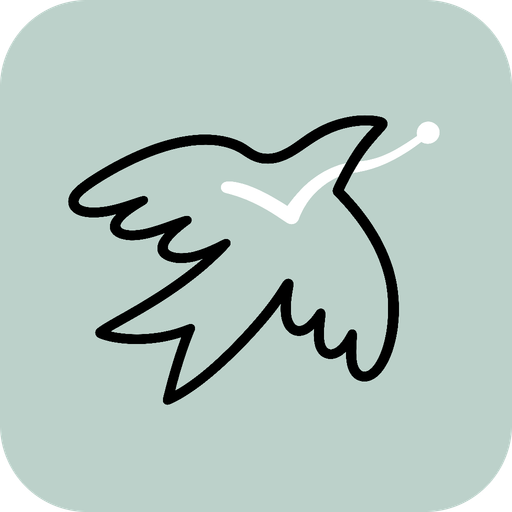}%
            }\ \textbf{Claude-Haiku-5.5}
            & 0.13 & 0.07 & 0.27 & 0.06 & 0.19 & 0.09 & 0.14 \\
            \quad w/ Domain Instruction
            & 0.45 & 0.49 & 0.49 & 0.29 & 0.50 & 0.04
            & 0.38 \gain{(+171.4\%)} \\
            \bottomrule
        \end{tabular}%
    }
    \caption{Task success rates across AutomationBench domains.
    Parentheses indicate relative changes in overall success rate.
    Component substitutions are compared with Claude-Opus-5.5;
    domain-instruction results are compared with the corresponding
    base model.}
    \label{tab:benchmark}
\end{table*}

\paragraph{Evaluation of LLM Agents' Tool Use.}
Existing benchmarks evaluate tool use at different levels. API-Bank assesses planning, retrieval, and API calling through executable dialogues \cite{li2023apibank}, while TaskBench evaluates task decomposition, tool selection, and parameter prediction through tool invocation graph prediction \cite{shen2024taskbench}. ToolRet focuses on retrieval from large tool collections \cite{shi2025toolret}, and BFCL evaluates function-call correctness across single-turn and multi-turn settings \cite{patil2025bfcl}. Interactive benchmarks assess tool use within broader workflows: AgentBench spans diverse environments \cite{liu2024agentbench}, AppWorld evaluates cross-application workflows through environment state changes \cite{trivedi2024appworld}, and ToolSandbox examines state dependencies and intermediate milestones \cite{lu2025toolsandbox}. The $\tau$-bench family evaluates policy adherence and reliable task completion in interactive settings \cite{yao2024taubench,barres2025tau2}, while AutomationBench tests business workflows requiring API discovery and cross-application coordination \cite{shepard2026automationbench}. Previous work typically evaluates either end-to-end task success or individual tool-use capabilities in isolation. Building on AutomationBench, PSE varies the LLM assigned to planning, selection, or execution while keeping the other components fixed across the same long-horizon tasks. By retaining final task success as the evaluation metric, PSE examines how component-level capability changes propagate through the agent loop, identifying potential bottlenecks and priorities for improvement.

\section{PSE Evaluation Protocol}
\label{sec:pse}

PSE decomposes each agent turn into three atomic capabilities: planning, selection, and execution. Each component can be assigned to a different LLM, enabling controlled evaluation of how component-level model substitutions affect final task success.

\paragraph{Three atomic capabilities.}
Let $h_t$ denote the interaction history at turn $t$ and $\mathcal{D}_t$ the cumulative registry of retrieved API documents, with $\mathcal{D}_0=\varnothing$. The planning model decides whether to search, execute, or finish:
\begin{equation}
z_t=f_P(h_t)\in\{\mathsf{search}(q_t),\mathsf{execute},\mathsf{finish}\}.
\end{equation}
A search decision invokes \texttt{search\_tools} with query $q_t$ and adds the returned documents to the registry. A finish decision terminates the rollout. An execute decision invokes the selection model to choose an API from the registry:
\begin{equation}
u_t=f_S(h_t,\mathcal{D}_t), \qquad u_t\in\mathrm{names}(\mathcal{D}_t).
\end{equation}
Given the selected API $u_t$ and its parameter schema $\sigma(u_t)$, the execution model generates the arguments:
\begin{equation}
\theta_t=f_E(h_t,u_t,\sigma(u_t)).
\end{equation}
The controller then invokes \texttt{execute\_tool} with $u_t$ and $\theta_t$ and appends the resulting observation to the history. Selection is restricted to retrieved APIs, and execution cannot change the selected API. Invalid outputs are recorded as protocol failures rather than repaired by another model.

\paragraph{Controlled evaluation.}
The control configuration assigns the reference model $M^\star$ to all three components. We then substitute a weaker model $m$ in one component at a time while keeping the others fixed:
\begin{align}
M^{\mathrm{ctl}} &= (M^\star,M^\star,M^\star),\\
M^{P\leftarrow m} &= (m,M^\star,M^\star),\\
M^{S\leftarrow m} &= (M^\star,m,M^\star),\\
M^{E\leftarrow m} &= (M^\star,M^\star,m).
\end{align}
All configurations are evaluated on the same tasks from identical initial environment states. We evaluate the native single-model agent separately to distinguish the effect of decomposition from that of component substitution. All substitutions are compared against the same decomposed control. Differences in final task success therefore capture the downstream effects of each component substitution, including error propagation and recovery.

\section{Experiments}

\subsection{Setup}
\label{sec:experimental-setup}

\paragraph{Benchmark and metric.}
We evaluate on six AutomationBench domains: Sales, Marketing, Operations, Support, Finance, and HR. Each domain contains 100 tasks, yielding 600 instances in total. We report \emph{Pass@1}, the fraction of tasks whose terminal environment state satisfies all task-specific verifier assertions. All configurations are evaluated on the same instances. Domain scores are averaged over 100 tasks, and the overall score is the micro-average across all 600 tasks. We run one rollout per task with a maximum of 50 logical agent turns. API calls execute in the benchmark's simulated environment, and success is determined by the terminal-state verifier.

\paragraph{PSE component controls.}
Following Section~\ref{sec:pse}, the control configuration assigns Claude-Opus-5.5 to all three components. We then replace planning, selection, or execution with Claude-Haiku-5.5 while retaining Opus for the other two components. All LLMs are configured to use adaptive thinking effort. Each configuration follows its own feedback-driven trajectory. Planning decides whether to search, execute, or finish; searches return up to 10 API documents; selection chooses from the cumulative registry of retrieved APIs; and execution generates the selected API's arguments. Thus, a component substitution can affect both the current action and subsequent decisions.

\paragraph{Domain-instruction condition.}
We additionally evaluate Opus and Haiku as single-model agents with a fixed domain API library and workflow instruction. For each domain $d$, the library is constructed as:
$\mathcal{L}_d=\bigcup_{x\in\mathcal{T}_d}\mathrm{APIs}(x)$
where $\mathcal{T}_d$ contains the 100 evaluation tasks and $\mathrm{APIs}(x)$ denotes the API set specified in task $x$'s specification. Claude-Opus-5.5 distills descriptions of workflows into a fixed instruction for each domain. Both models receive the same library and instruction at the start of each rollout and invoke APIs directly without \texttt{search\_tools}. This condition is designed to reduce agents' unfamiliarity with domain-specific workflows and mitigate the effects of limited tool planning capabilities. No test-set labels are leaked.

\subsection{Results and Discussion}

\paragraph{\textcolor{langdarkblue}{Tool planning is most sensitive to model capability.}}
Table~\ref{tab:benchmark} shows that substituting Haiku for Opus in any PSE component reduces overall task success. Planning exhibits the largest decline: Pass@1 decreases from 0.29 to 0.15, a relative reduction of 48.3\%, compared with 20.7\% for selection and 27.6\% for execution. These results identify planning as the main bottleneck under the evaluated model substitutions. Planning governs whether the agent searches for additional tools, executes an available API, or terminates the task. An inappropriate decision can therefore redirect the workflow and affect subsequent tool choices and observations, even when selection and execution remain assigned to the stronger model. Planning also requires adapting to feedback: when an action fails or produces an unexpected result, the agent must reassess the situation and revise its next step. Effective tool planning thus involves both guiding the workflow and re-planning in response to problems encountered during execution.

\paragraph{\textcolor{langdarkblue}{Limited domain knowledge and experience may constrain tool planning.}}
In zero-shot settings, stronger LLMs may be better equipped to navigate unfamiliar workflows. However, the domain-instruction condition substantially improves task success for both models, increasing overall Pass@1 from 0.29 to 0.55 for Opus and from 0.05 to 0.38 for Haiku. Both models improve across all six domains, although the magnitude varies considerably. Notably, Haiku under this condition outperforms the Opus baseline overall. This suggests that providing domain tools and descriptions of successful workflows can help agents navigate unfamiliar procedures. Such guidance makes relevant operations and their dependencies more explicit, reducing the need to infer workflows from tool discovery and observations alone and easing the demands on tool planning. These results suggest that domain guidance is a low-hanging fruit for agent improvement: in our experiments, its gains exceed those obtained by switching from Haiku to Opus under the baseline condition.

\paragraph{\textcolor{langdarkblue}{Improving tool planning is a priority for agent development.}}
Together, these findings suggest prioritizing tool planning and domain-specific workflow knowledge when developing long-horizon agents. Improving API selection or argument generation remains important, but their benefits may be constrained when the agent pursues an inappropriate action or terminates prematurely. In summary, many existing approaches improve tool use by refining planning structures and modeling task dependencies \cite{shen2023hugginggpt,song2023restgpt,xu2023rewoo,kim2024llmcompiler,chen2026gtool,yang2026tooltree}. Our controlled evaluation provides empirical support for this direction by showing how reduced planning capability affects final task success. These findings motivate the development of agents that better acquire domain-specific workflow knowledge and learn in-domain planning strategies.

\section{Conclusion}
We introduced PSE, a controlled evaluation framework that decomposes long-horizon tool use into three atomic capabilities: planning, selection, and execution. By substituting a weaker LLM in one component at a time while keeping the others fixed, we examined how component-level changes affect final task success. Experiments across six domains identified planning as the most sensitive component to model substitution, while domain instructions substantially improved performance. These findings highlight improving tool planning as priorities for future LLM agent development.

\section*{Limitations}
Our evaluation covers a limited set of representative use cases across six domains and may not capture the full complexity of real-world tasks and environments. Future work can extend this analysis to a broader range of real-world scenarios. \textit{Lang has many limitations, but Bing should also handle the future work.}

\section*{Ethical Considerations}
Tool-using agents can make consequential changes to external systems. They should therefore be evaluated in controlled environments and deployed with appropriate permissions, safeguards, and human oversight.


\bibliography{custom}

@inproceedings{yao2023react,
  title     = {{ReAct}: Synergizing Reasoning and Acting in Language Models},
  author    = {Yao, Shunyu and Zhao, Jeffrey and Yu, Dian and Du, Nan
               and Shafran, Izhak and Narasimhan, Karthik and Cao, Yuan},
  booktitle = {International Conference on Learning Representations},
  year      = {2023},
  url       = {https://openreview.net/forum?id=WE_vluYUL-X}
}

@inproceedings{schick2023toolformer,
  title     = {Toolformer: Language Models Can Teach Themselves to Use Tools},
  author    = {Schick, Timo and Dwivedi-Yu, Jane and Dessi, Roberto
               and Raileanu, Roberta and Lomeli, Maria and Hambro, Eric
               and Zettlemoyer, Luke and Cancedda, Nicola and Scialom, Thomas},
  booktitle = {Advances in Neural Information Processing Systems},
  volume    = {36},
  pages     = {68539--68551},
  year      = {2023},
  doi       = {10.52202/075280-2997},
  url       = {https://proceedings.neurips.cc/paper_files/paper/2023/hash/d842425e4bf79ba039352da0f658a906-Abstract-Conference.html}
}

@inproceedings{patil2024gorilla,
  title     = {Gorilla: Large Language Model Connected with Massive {APIs}},
  author    = {Patil, Shishir G. and Zhang, Tianjun and Wang, Xin
               and Gonzalez, Joseph E.},
  booktitle = {Advances in Neural Information Processing Systems},
  volume    = {37},
  year      = {2024},
  doi       = {10.52202/079017-4020},
  url       = {https://proceedings.neurips.cc/paper_files/paper/2024/hash/e4c61f578ff07830f5c37378dd3ecb0d-Abstract-Conference.html}
}

@inproceedings{qin2024toolllm,
  title     = {{ToolLLM}: Facilitating Large Language Models to Master
               16000+ Real-world {APIs}},
  author    = {Qin, Yujia and Liang, Shihao and Ye, Yining and Zhu, Kunlun
               and Yan, Lan and Lu, Yaxi and Lin, Yankai and Cong, Xin
               and Tang, Xiangru and Qian, Bill and Zhao, Sihan
               and Hong, Lauren and Tian, Runchu and Xie, Ruobing
               and Zhou, Jie and Gerstein, Mark and Li, Dahai
               and Liu, Zhiyuan and Sun, Maosong},
  booktitle = {International Conference on Learning Representations},
  year      = {2024},
  url       = {https://proceedings.iclr.cc/paper_files/paper/2024/hash/28e50ee5b72e90b50e7196fde8ea260e-Abstract-Conference.html}
}

@inproceedings{shen2023hugginggpt,
  title     = {{HuggingGPT}: Solving {AI} Tasks with {ChatGPT}
               and its Friends in Hugging Face},
  author    = {Shen, Yongliang and Song, Kaitao and Tan, Xu
               and Li, Dongsheng and Lu, Weiming and Zhuang, Yueting},
  booktitle = {Advances in Neural Information Processing Systems},
  volume    = {36},
  year      = {2023},
  doi       = {10.52202/075280-1657},
  url       = {https://proceedings.neurips.cc/paper_files/paper/2023/hash/77c33e6a367922d003ff102ffb92b658-Abstract-Conference.html}
}

@article{xu2023rewoo,
  title   = {{ReWOO}: Decoupling Reasoning from Observations
             for Efficient Augmented Language Models},
  author  = {Xu, Binfeng and Peng, Zhiyuan and Lei, Bowen
             and Mukherjee, Subhabrata and Liu, Yuchen and Xu, Dongkuan},
  journal = {arXiv preprint arXiv:2305.18323},
  year    = {2023},
  doi     = {10.48550/arXiv.2305.18323},
  url     = {https://arxiv.org/abs/2305.18323}
}

@inproceedings{kim2024llmcompiler,
  title     = {An {LLM} Compiler for Parallel Function Calling},
  author    = {Kim, Sehoon and Moon, Suhong and Tabrizi, Ryan
               and Lee, Nicholas and Mahoney, Michael W.
               and Keutzer, Kurt and Gholami, Amir},
  booktitle = {Proceedings of the 41st International Conference
               on Machine Learning},
  series    = {Proceedings of Machine Learning Research},
  volume    = {235},
  pages     = {24370--24391},
  publisher = {PMLR},
  year      = {2024},
  url       = {https://proceedings.mlr.press/v235/kim24y.html}
}

@inproceedings{li2023apibank,
  title     = {{API}-Bank: A Comprehensive Benchmark for Tool-Augmented {LLM}s},
  author    = {Li, Minghao and Zhao, Yingxiu and Yu, Bowen
               and Song, Feifan and Li, Hangyu and Yu, Haiyang
               and Li, Zhoujun and Huang, Fei and Li, Yongbin},
  booktitle = {Proceedings of the 2023 Conference on Empirical
               Methods in Natural Language Processing},
  pages     = {3102--3116},
  publisher = {Association for Computational Linguistics},
  year      = {2023},
  doi       = {10.18653/v1/2023.emnlp-main.187},
  url       = {https://aclanthology.org/2023.emnlp-main.187/}
}

@inproceedings{shen2024taskbench,
  title     = {{TaskBench}: Benchmarking Large Language Models
               for Task Automation},
  author    = {Shen, Yongliang and Song, Kaitao and Tan, Xu
               and Zhang, Wenqi and Ren, Kan and Yuan, Siyu
               and Lu, Weiming and Li, Dongsheng and Zhuang, Yueting},
  booktitle = {Advances in Neural Information Processing Systems},
  volume    = {37},
  pages     = {4540--4574},
  year      = {2024},
  doi       = {10.52202/079017-0148},
  url       = {https://proceedings.neurips.cc/paper_files/paper/2024/hash/085185ea97db31ae6dcac7497616fd3e-Abstract-Datasets_and_Benchmarks_Track.html}
}

@inproceedings{shi2025toolret,
  title     = {Retrieval Models Aren't Tool-Savvy: Benchmarking Tool
               Retrieval for Large Language Models},
  author    = {Shi, Zhengliang and Wang, Yuhan and Yan, Lingyong
               and Ren, Pengjie and Wang, Shuaiqiang
               and Yin, Dawei and Ren, Zhaochun},
  booktitle = {Findings of the Association for Computational Linguistics:
               ACL 2025},
  pages     = {24497--24524},
  publisher = {Association for Computational Linguistics},
  year      = {2025},
  doi       = {10.18653/v1/2025.findings-acl.1258},
  url       = {https://aclanthology.org/2025.findings-acl.1258/}
}

@inproceedings{patil2025bfcl,
  title     = {The Berkeley Function Calling Leaderboard ({BFCL}):
               From Tool Use to Agentic Evaluation of Large Language Models},
  author    = {Patil, Shishir G and Mao, Huanzhi and Yan, Fanjia
               and Ji, Charlie Cheng-Jie and Suresh, Vishnu
               and Stoica, Ion and Gonzalez, Joseph E.},
  booktitle = {Proceedings of the 42nd International Conference
               on Machine Learning},
  series    = {Proceedings of Machine Learning Research},
  volume    = {267},
  pages     = {48371--48392},
  publisher = {PMLR},
  year      = {2025},
  url       = {https://proceedings.mlr.press/v267/patil25a.html}
}

@inproceedings{liu2024agentbench,
  title     = {{AgentBench}: Evaluating {LLM}s as Agents},
  author    = {Liu, Xiao and Yu, Hao and Zhang, Hanchen and Xu, Yifan
               and Lei, Xuanyu and Lai, Hanyu and Gu, Yu
               and Ding, Hangliang and Men, Kaiwen and Yang, Kejuan
               and Zhang, Shudan and Deng, Xiang and Zeng, Aohan
               and Du, Zhengxiao and Zhang, Chenhui and Shen, Sheng
               and Zhang, Tianjun and Su, Yu and Sun, Huan
               and Huang, Minlie and Dong, Yuxiao and Tang, Jie},
  booktitle = {International Conference on Learning Representations},
  year      = {2024},
  url       = {https://proceedings.iclr.cc/paper_files/paper/2024/hash/e9df36b21ff4ee211a8b71ee8b7e9f57-Abstract-Conference.html}
}

@inproceedings{trivedi2024appworld,
  title     = {{AppWorld}: A Controllable World of Apps and People
               for Benchmarking Interactive Coding Agents},
  author    = {Trivedi, Harsh and Khot, Tushar and Hartmann, Mareike
               and Manku, Ruskin and Dong, Vinty and Li, Edward
               and Gupta, Shashank and Sabharwal, Ashish
               and Balasubramanian, Niranjan},
  booktitle = {Proceedings of the 62nd Annual Meeting of the Association
               for Computational Linguistics (Volume 1: Long Papers)},
  pages     = {16022--16076},
  publisher = {Association for Computational Linguistics},
  year      = {2024},
  doi       = {10.18653/v1/2024.acl-long.850},
  url       = {https://aclanthology.org/2024.acl-long.850/}
}

@inproceedings{lu2025toolsandbox,
  title     = {{ToolSandbox}: A Stateful, Conversational, Interactive
               Evaluation Benchmark for {LLM} Tool Use Capabilities},
  author    = {Lu, Jiarui and Holleis, Thomas and Zhang, Yizhe
               and Aumayer, Bernhard and Nan, Feng and Bai, Haoping
               and Ma, Shuang and Ma, Shen and Li, Mengyu
               and Yin, Guoli and Wang, Zirui and Pang, Ruoming},
  booktitle = {Findings of the Association for Computational Linguistics:
               NAACL 2025},
  pages     = {1160--1183},
  publisher = {Association for Computational Linguistics},
  year      = {2025},
  doi       = {10.18653/v1/2025.findings-naacl.65},
  url       = {https://aclanthology.org/2025.findings-naacl.65/}
}

@article{yao2024taubench,
  title   = {{$\tau$-bench}: A Benchmark for Tool-Agent-User
             Interaction in Real-World Domains},
  author  = {Yao, Shunyu and Shinn, Noah and Razavi, Pedram
             and Narasimhan, Karthik},
  journal = {arXiv preprint arXiv:2406.12045},
  year    = {2024},
  doi     = {10.48550/arXiv.2406.12045},
  url     = {https://arxiv.org/abs/2406.12045}
}

@article{barres2025tau2,
  title   = {{$\tau^2$-Bench}: Evaluating Conversational Agents
             in a Dual-Control Environment},
  author  = {Barres, Victor and Dong, Honghua and Ray, Soham
             and Si, Xujie and Narasimhan, Karthik},
  journal = {arXiv preprint arXiv:2506.07982},
  year    = {2025},
  doi     = {10.48550/arXiv.2506.07982},
  url     = {https://arxiv.org/abs/2506.07982}
}

@article{shepard2026automationbench,
  title   = {{AutomationBench}},
  author  = {Shepard, Daniel and Salimans, Robin},
  journal = {arXiv preprint arXiv:2604.18934},
  year    = {2026},
  doi     = {10.48550/arXiv.2604.18934},
  url     = {https://arxiv.org/abs/2604.18934}
}

@article{song2023restgpt,
  title   = {{RestGPT}: Connecting Large Language Models with Real-World {RESTful} {APIs}},
  author  = {Song, Yifan and Xiong, Weimin and Zhu, Dawei and
             Wu, Wenhao and Qian, Han and Song, Mingbo and
             Huang, Hailiang and Li, Cheng and Wang, Ke and
             Yao, Rong and Tian, Ye and Li, Sujian},
  journal = {arXiv preprint arXiv:2306.06624},
  year    = {2023},
  url     = {https://arxiv.org/abs/2306.06624}
}

@inproceedings{chen2026gtool,
  title     = {{GTool}: Graph Enhanced Tool Planning with Large Language Model},
  author    = {Chen, Wenjie and Yao, Di and Li, Wenbin and
               Meng, Xuying and Gong, Chang and Bi, Jingping},
  booktitle = {International Conference on Learning Representations},
  year      = {2026},
  url       = {https://proceedings.iclr.cc/paper_files/paper/2026/hash/715cf37ace0cb884530ab593eea6a50c-Abstract-Conference.html}
}

@article{yang2026tooltree,
  title   = {{ToolTree}: Efficient {LLM} Agent Tool Planning via
             Dual-Feedback Monte Carlo Tree Search and Bidirectional Pruning},
  author  = {Yang, Shuo and Han, Soyeon Caren and Ding, Yihao and
             Wang, Shuhe and Hoy, Eduard},
  journal = {arXiv preprint arXiv:2603.12740},
  year    = {2026},
  url     = {https://arxiv.org/abs/2603.12740}
}





\end{document}